%% file: main.tex
\documentclass[runningheads]{llncs}
\usepackage{url}
\usepackage[T1]{fontenc}
\usepackage{graphicx}
\usepackage{amsmath}
\usepackage{amssymb}
\usepackage{booktabs}
\usepackage{cleveref}
\usepackage{bbding}

\newcommand{\eg}{\emph{e.g.}\@}

\begin{document}
\title{Align Where the Words Look: Cross-Attention-Guided Patch Alignment with Contrastive and Transport Regularization for Bengali Captioning}
\titlerunning{Align Where the Words Look}
%

\author{
Riad Ahmed Anonto\inst{1}\orcidID{0009-0008-4243-9990}\textsuperscript{*} \and
Sardar Md. Saffat Zabin\inst{1}\orcidID{0009-0004-1884-8269}\textsuperscript{*,\Envelope} \and
M. Saifur Rahman\inst{1}\orcidID{0000-0002-9887-4456}
}

\authorrunning{R.A. Anonto et al.}

\institute{Bangladesh University of Engineering and Technology (BUET), Dhaka, Bangladesh\\
\email{riadahmedanonto355@gmail.com},
\email{saffatzabin08430843@gmail.com},
\email{mrahman@cse.buet.ac.bd}}

%
\maketitle              
\stepcounter{footnote}
\footnotetext{Code: \url{https://github.com/Anonto050/Bn-PAL-Cap}. 
The synthetic dataset is available from the corresponding author upon reasonable request.}
\stepcounter{footnote}
\footnotetext{\textsuperscript{*} These authors contributed equally to this work.}

\input{sec/0_abstract}    
\input{sec/1_intro}
\input{sec/related_works}

\input{sec/architecture}
\input{sec/methodology}

\input{sec/experiments}

\input{sec/llms}
\input{sec/conclusion}
\subsubsection{Disclosure of Interests.}
The authors have no competing interests to declare that are relevant to the
content of this article.
\bibliographystyle{splncs04}
\bibliography{main}
%




\end{document}

%% file: sec/0_abstract.tex
\begin{abstract}
Adapting vision–language models in low-resource languages remains challenging, as they often produce fluent text about the wrong objects. This stems from scarce paired data, translation pivots that break alignment, and English-centric pretraining that ignores target-language semantics. We address this with a compute-aware Bengali captioning pipeline trained on LaBSE-verified EN–BN pairs and 110k bilingual-prompted synthetic images. A frozen MaxViT yields stable visual patches, a Bengali-native mBART-50 decodes, and a lightweight bridge links the modalities. Our core novelty is a tri-loss objective: Patch-Alignment Loss (PAL) aligns real and synthetic patch descriptors using decoder cross-attention, InfoNCE enforces global real–synthetic separation, and Sinkhorn-based OT ensures balanced fine-grained patch correspondence. This PAL+InfoNCE+OT synergy tightens real-synthetic patch alignment, reduces spurious matches, and drives strong gains on Flickr30k-1k (BLEU-4 12.29, METEOR 27.98, BERTScore-F1 71.20) and MSCOCO-1k (BLEU-4 12.00, METEOR 28.14, BERTScore-F1 75.40), outperforming strong CE baselines and narrowing the real–synthetic centroid gap by 41\%.

\keywords{Bengali image captioning \and Low-resource vision-language learning \and Patch alignment \and Cross-attention \and Synthetic data \and Optimal transport}
\end{abstract}

%% file: sec/1_intro.tex
\section{Introduction}
\label{sec:intro}

\noindent
Image captioning has advanced rapidly on English benchmarks, powered by large-scale vision–language pretraining and abundant web alt-text (\eg, CLIP, BLIP, SimVLM~\cite{li2022blip,radford2021clip,wang2022simvlm}). Pipelines overwhelmingly privilege English and a few high-resource languages. For Bengali—the world’s 7\textsuperscript{th} most spoken—paired image–text data are scarce, and available sets are small or narrow. Multilingual captioners often pivot through English or weakly aligned alt-text \cite{unpaired_pivot,wu-etal-2023-cross2stra,Zhou2021UC2:}, which broadens vocabulary but often severs the link between caption tokens and image content, producing fluent text about the wrong objects. English-centric pretraining also lacks patch-level, target-language-conditioned alignment.

Recent work has tried to strengthen the text-image link from two directions. ViPCap~\cite{Kim_Lee_Kim_Kim_2025} enriches visual features by retrieving text captions, converting them into semantic visual prompts, and aligning them with image patches. SynTIC~\cite{Liu_Liu_Ma_2024} improves cross-modal alignment by generating synthetic images for text-only captions, refining pseudo image features with contrastive loss, and projecting them into text space. While effective on rich English benchmarks, both approaches overlook two challenges crucial for Bengali: ViPCap depends on retrieved English text and does not condition alignment on target-language tokens, while SynTIC never uses real images during training and lacks patch-level alignment, limiting generalization.

We address these gaps with a compute-aware Bengali captioning pipeline. A frozen MaxViT~\cite{tu2022maxvit} backbone supplies stable visual patches, a Bengali-native mBART-50~\cite{liu2020mbart} decoder generates text, and a lightweight linear+LayerNorm bridge links the modalities. We translate and verify EN–BN caption pairs using LaBSE~\cite{feng-etal-2022-language}, render paired synthetic images via bilingual prompts (``A photo of: EN. In Bengali: BN'') using Kandinsky~2.1~\cite{kandinsky21}, and train on real images with cross-entropy plus a three-part alignment objective. This objective combines: (i) a Patch-Alignment Loss (PAL) that reuses decoder cross-attention to form text-conditioned weights over patches and align pooled real and synthetic descriptors, (ii) an Information Noise-Contrastive Estimation (InfoNCE) \cite{oord2018representation} term that enforces global discrimination between real and synthetic pooled features, and (iii) a Sinkhorn-based Optimal Transport (OT) \cite{cuturi2013sinkhorn} term that enforces balanced patch-wise matching under the same PAL attention. PAL steers learning to caption-relevant regions, while InfoNCE pulls apart background-dominated embeddings and OT spreads alignment mass across multiple fine-grained patches—together forming a mutually reinforcing cycle that neither achieves alone.

Synthetic images expand visual diversity without requiring scarce Bengali-paired data, while bilingual prompts inject target-language signal directly into the vision pathway. Freezing the vision tower preserves stable image features, and training the decoder alone ensures native Bengali fluency.

Our contributions are as follows:
\begin{itemize}
  \item We propose a novel tri-loss framework combining Patch-Alignment Loss (PAL), InfoNCE, and Sinkhorn-based OT, where PAL aligns decoder cross-attended real vs.\ synthetic patches, InfoNCE enhances global discriminability, and OT enforces mass-balanced fine-grained patch correspondence.
  \item We construct a Bengali captioning corpus and synthesis pipeline by verifying EN–BN caption pairs and generating large-scale bilingual-prompted synthetic images to form real–synthetic training triplets.
  \item We design a lightweight training stack where a frozen MaxViT vision backbone and an mBART-50 Bengali decoder are connected by a linear+LayerNorm bridge, enabling efficient training on a single GPU and easy transfer to other low-resource languages.
  \item We show that this combined PAL+InfoNCE+OT objective outperforms a CE-only baseline on MSCOCO-1k and Flickr30k-1k~\cite{lin2014coco,young-etal-2014-image}, boosting BLEU-$n$, METEOR, and BERTScore while yielding more stable captions across domains.
\end{itemize}

%% file: sec/related_works.tex
\section{Related Work}
\label{sec:related}

\paragraph{Bengali image captioning.}
Early Bangla/Bengali captioners pair CNN encoders with RNN/Transformer decoders trained on small, task-specific corpora~\cite{bornon2021bengali,Humaira2021A,Shah2021Bornon:}.
BAN-Cap expands data but underscores the scarcity of high-quality, domain-diverse supervision~\cite{hossain2022bancap}.
Prior systems often (i) fine-tune English-centric vision backbones end-to-end on tiny sets (overfitting), or (ii) pivot via machine translation (MT) at train/test time (hurting fluency/morphology).
We instead freeze a strong encoder, adapt a Bengali-native decoder, and enforce text-conditioned alignment between real and synthetic imagery only where the caption attends.

\paragraph{Low-resource \& multilingual captioning.}
Few-/low-shot transfer to new languages has been explored~\cite{Khanuja2021MuRILMR,Nath2022Image,rajpal2023lmcap}; Indian-language pipelines frequently translate references or pseudo-labels~\cite{patil2020robust}, improving coverage but rarely addressing where alignment occurs.
Our design avoids translation at decoding (mBART-50 tokenizer) and injects a direct patch-level alignment signal via PAL.
Orthogonally, test-time caption–image consistency (ICE) improves out of domain (OOD) robustness without retraining~\cite{zeroshot}; we reshape the interface during training.

\paragraph{General-purpose captioners \& V+L pretraining.}
Advances in English captioning span attention and large-scale pretraining—AoANet~\cite{huang2019aoanet}, SimVLM~\cite{wang2022simvlm}, BLIP~\cite{li2022blip}, UNITER~\cite{chen2020uniter}.
These excel with abundant pairs but do not align real-synthetic images conditioned on in-loop decoding attention for a non-English captioner.
Gaze-estimation signals can guide attention~\cite{gazing}; we instead exploit the model’s own cross-attention.

\paragraph{Patch-level alignment \& synthetic data.}
PatchNCE aligns patches without language~\cite{park2020cut}, and OT-style word--region matching appears in pretraining~\cite{chen2020uniter,cuturi2013sinkhorn} but not inside the caption loop.
Synthetic augmentation aids specialized captioning (e.g., BLIP2IDC for image-difference)~\cite{reframe}, yet typically lacks caption-conditioned real--synthetic alignment.
ViPCap~\cite{Kim_Lee_Kim_Kim_2025} retrieves English captions as prompts to enrich visual features, while SynTIC~\cite{Liu_Liu_Ma_2024} contrasts synthetic features against text-only captions without using real images during training.
Both methods improve alignment on high-resource English benchmarks, but ViPCap depends on English text pivots that break weak alignment in Bengali, and SynTIC lacks patch-level alignment and fails to leverage real visual context.

Our approach embeds alignment directly into decoding: PAL reuses decoder cross-attention to pool caption-relevant patches and aligns pooled real--synthetic descriptors with a cosine loss, while InfoNCE enforces global discrimination and OT sharpens fine-grained correspondence.
This joint PAL+InfoNCE+OT design unifies caption-time relevance with multi-scale alignment, conditioning cross-modal alignment directly on Bengali tokens.

%% file: sec/architecture.tex
\section{Architecture Overview}
\label{sec:arch}

\subsection{Vision encoder — MaxViT (Frozen)}
\label{ssec:maxvit}

MaxViT is a hierarchical vision transformer that interleaves lightweight MBConv (inverted residual convolution) blocks with multi-axis self-attention (Max-SA). Max-SA factorizes dense global attention into two sparse, complementary views: window attention for local self-attention inside fixed windows, and grid attention for strided global token exchange. Alternating these views provides an effective global receptive field while keeping complexity near-linear in the number of patches. The final stages retain relatively high spatial resolution, which is useful for patch-level reasoning.

We use MaxViT because its large-scale pretraining yields strong, transferable features that serve as stable visual priors when Bengali supervision is scarce. The hierarchy exposes both final and penultimate feature maps, which can be projected into the decoder space for multi-scale alignment. These dense yet semantically structured patch features are ideal for PAL, where local relevance must be isolated, and they also benefit InfoNCE and OT: their wide variation across samples provides hard negatives for contrastive discrimination, and their clean spatial layout makes cross-sample patch matching via OT well-conditioned. Freezing the encoder preserves this structure, reducing trainable parameters and preventing catastrophic forgetting.

\subsection{Language decoder — mBART-50 (Trainable)}
\label{ssec:mbart}

mBART-50 is a sequence-to-sequence transformer pretrained by multilingual denoising; the \texttt{bn\_IN} variant provides native Bengali subword tokenization and strong priors over morphology and word order. The decoder consumes projected visual tokens via cross-attention and autoregressively predicts Bengali tokens under teacher forcing. Its cross-attention maps also indicate which patches support each token, which we reuse to weight patches in PAL.

We fine-tune only the decoder, applying gradient checkpointing for memory efficiency and enforcing \texttt{bn\_IN} as the BOS token. Its built-in multilingual priors support fluent generation from sparse supervision, and its stable cross-attention activations give consistent token–patch maps for PAL. These same activations provide coherent global embeddings for InfoNCE and well-localized patch descriptors for OT, making the decoder a natural fit for all three objectives.

%% file: sec/methodology.tex
\section{Methodology}
\label{sec:method}

\begin{figure*}[t]
  \centering
  \includegraphics[width=0.98\textwidth]{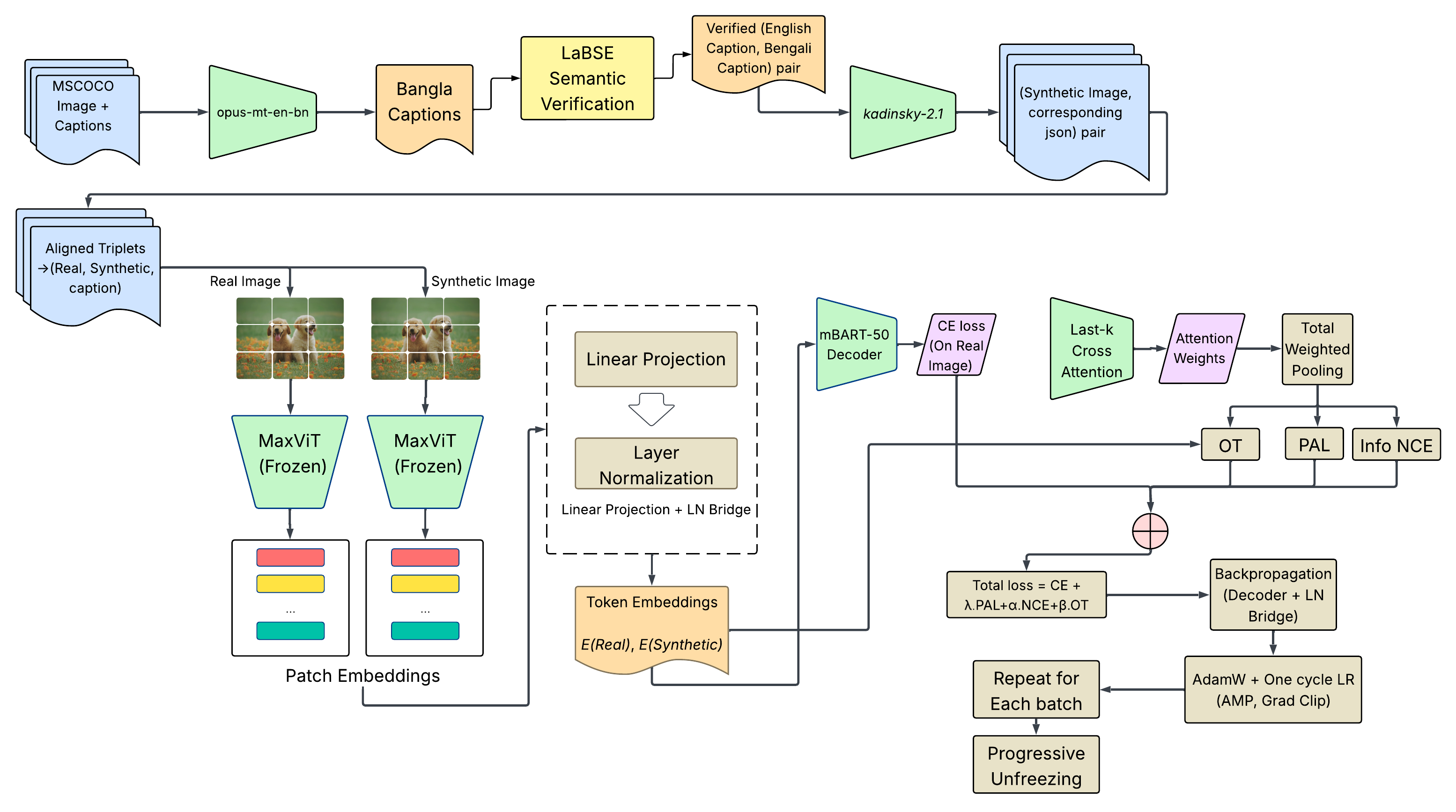}
  \caption{Pipeline. MSCOCO captions are translated to Bengali and LaBSE-verified, then used to render a synthetic image per caption, forming (real, synthetic, BN caption) triplets. A frozen MaxViT feeds a Linear+LN bridge into an mBART-50 decoder. Last-$k$ cross-attention provides text-conditioned patch weights for pooling. Training uses CE on real images and PAL on pooled real–synthetic descriptors, with InfoNCE and Sinkhorn OT as lightweight regularizers; only the bridge and decoder are updated.}
  \label{fig:overview}
\end{figure*}

Our pipeline (Fig.~\ref{fig:overview}) assembles a Bengali-aligned corpus and trains a captioner with three complementary objectives: cross-entropy (CE) on real images for fluent generation; a Patch-Alignment Loss (PAL) that uses decoder cross-attention to weight caption-relevant patches and align their real–synthetic descriptors; and two auxiliaries—InfoNCE for global discrimination and Sinkhorn-based OT for fine-grained patch correspondence. PAL localizes what to align, InfoNCE sharpens inter-sample separation, and OT regularizes local structure; together they improve real-synthetic alignment under scarce Bengali supervision. 

\subsection{Problem Setup \& Notation}
\label{ssec:problem-notation}

We study Bengali image captioning under limited paired supervision.  
Given an image $x\!\in\!\mathbb{R}^{3\times H_0\times W_0}$ and a Bengali caption $y=(y_{1},\ldots,y_{T})$, a vision–language model with parameters $\theta$ predicts the most likely sequence
\begin{equation}
\hat{y}=\arg\max_{y}\; p_{\theta}(y\,|\,x).
\end{equation}

The training corpus combines two aligned sources:  
(i) a \emph{real} set $D_r=\{(x_i,y_i)\}_{i=1}^{N_r}$ containing MSCOCO images with LaBSE-verified Bengali translations, and  
(ii) a \emph{synthetic} set $D_s=\{(\tilde{x}_i,y_i)\}_{i=1}^{N_s}$ where each $\tilde{x}_i$ is rendered from the same caption $y_i$ using bilingual prompts.  
Unless stated otherwise, indices in $D_s$ are matched to $D_r$, forming triplets $(x_i,\tilde{x}_i,y_i)$ that share captions but differ visually, giving diverse positive pairs.

A frozen vision encoder $F(\cdot)$ (MaxViT; see Sec.~\ref{sec:arch}) maps an image to a final-stage feature map
\begin{equation}
F(x)\in\mathbb{R}^{C\times H\times W},\qquad S=H\cdot W.
\end{equation}
A linear projection followed by LayerNorm converts these spatial features into patch tokens for the decoder:
\begin{equation}
E(x)=\mathrm{LN}\!\big(\mathrm{reshape}(F(x))^\top W_p\big)\in\mathbb{R}^{S\times D},
\end{equation}
where $W_p\!\in\!\mathbb{R}^{C\times D}$ and $D$ is the decoder hidden size.  
When available, a penultimate feature stage $F^{(p)}(x)$ is similarly projected to $E^{(p)}(x)\!\in\!\mathbb{R}^{S^{(p)}\times D}$, enabling optional multi-scale alignment.

\subsection{Building a Bengali-Aligned Training Set: Translation $\rightarrow$ Verification $\rightarrow$ Synthesis}
\label{ssec:data-pipeline}

\noindent Our objective is to construct a Bengali-aligned multimodal set $\mathcal{D}=\{(x_i, y_i^{\text{bn}})\}$ from MSCOCO English captions and a synthetic companion $\tilde{\mathcal{D}}=\{(\tilde{x}_i, y_i^{\text{bn}})\}$ where each $\tilde{x}_i$ is generated from the same caption $y_i^{\text{bn}}$. This paired design enables controlled real–synthetic alignment for PAL (\S\ref{ssec:pal}).

\subsubsection{Preprocessing and Translation}
We begin with MSCOCO’s 414{,}113 English captions over 82{,}783 images. From the official JSON we keep \texttt{caption\_id}, \texttt{image\_id}, and the English caption $y_i^{\text{en}}$. Captions are translated to Bengali with the transformer model \texttt{helsinki/opus-mt-en-bn} in batches of 128 for memory efficiency, yielding $y_i^{\text{bn}}$. The translated text is stored alongside the original, preserving one-to-many captioning per image.

\subsubsection{Semantic Verification (Security Layer)}
To ensure translations preserve meaning, we compute cross-lingual sentence embeddings with LaBSE and measure cosine similarity
\begin{equation}
s_i=\cos\big(\varphi(y_i^{\text{en}}),\ \varphi(y_i^{\text{bn}})\big).
\end{equation}
A pair is accepted iff $s_i \ge \tau$ with $\tau=0.55$; otherwise it is discarded. MSCOCO is processed in 10 shards (translate $\rightarrow$ verify independently); per-shard CSVs (with similarity and a validity flag) are later merged. This acts as a semantic integrity filter that removes mistranslations and off-topic outputs.
Through manual inspection, we found captions above this threshold to be semantically aligned; setting $\tau$ higher pruned too many valid pairs and hurt data coverage for the captioner. Using $\tau{=}0.55$ yielded $\sim$330{,}000 selected captions overall.

\subsubsection{Synthetic Image Generation with Joint Prompts}
For each verified pair $(y_i^{\text{en}},y_i^{\text{bn}})$ we synthesize an image $\tilde{x}_i$ using \emph{Kandinsky-2.1} (prior+decoder) under a bilingual prompt:
\begin{equation}
\texttt{"A photo of: } y_i^{\text{en}} \texttt{. In Bengali: } y_i^{\text{bn}}\texttt{"}.
\end{equation}
Because Kandinsky enforces a $\sim$77-token limit, we apply token-aware truncation that prioritizes content words in both languages ($\approx 37$ English $+$ $\approx 40$ Bengali tokens). Images are generated at $768{\times}768$ in \texttt{fp16} on GPU with a fixed seed (42) and a conservative negative prompt (e.g., ``low quality, bad anatomy, watermark'') to reduce artifacts. Within our compute budget, we synthesized 110{,}000 images in total; all runs preserve prompts, seeds, and hashes for reproducibility.

\noindent Each synthesized image is accompanied by a JSON sidecar storing the joint prompt, model versions, and a SHA-256 hash of the prompt, enabling audits and exact regeneration.

\paragraph{Why synthetic Images (and Joint Bilingual Prompts)?}
Bengali captioning lacks large, diverse pairs; synthesis augments each caption $y$ with multiple visual realizations $\tilde{x}\!\sim\!p(\tilde{x}\mid y)$, improving long-tail coverage without extra annotation. Because $\tilde{x}$ is conditioned on the \emph{same} $y$, the label remains consistent while styles/layouts vary, regularizing caption conditioned visual representation. We supervise language only on real $(x,y)$ (CE) and use $\tilde{x}$ solely via PAL to align text-attended regions, mitigating distribution shift from generator artifacts. Including Bengali in the prompt injects Bengali signal into the text-to-vision path and strengthens cross-lingual alignment---preferred when the goal is Bengali--vision alignment; pure English prompts are only preferable if image fidelity alone is the objective.

\subsection{Training objectives and Data Mixing}
\label{ssec:objectives}

We optimize a joint objective that couples caption learning on real images with caption-conditioned alignment between real–synthetic pairs:

\begin{equation}
\label{eq:joint-loss}
\begin{aligned}
\mathcal{L} &= \mathcal{L}_{\text{CE}}(x,y)
 + \lambda_{\text{PAL}}\,\mathcal{L}_{\text{PAL}}(x,\tilde{x},y) \\
&\quad + \alpha\,\mathcal{L}_{\text{InfoNCE}}
 + \beta\,\mathcal{L}_{\text{OT}} .
\end{aligned}
\end{equation}

Here, $\mathcal{L}_{\text{CE}}$ is the negative log-likelihood on real $(x,y)$; $\mathcal{L}_{\text{PAL}}$ pools decoder–cross-attended patches to align real vs.\ synthetic evidence for the same caption $y$; $\mathcal{L}_{\text{InfoNCE}}$ contrasts pooled real/synthetic descriptors across the batch; and $\mathcal{L}_{\text{OT}}$ applies entropic Sinkhorn transport over top-$k$ weighted patches with a cosine cost. In our main model $\lambda_{\text{PAL}},\alpha,\beta>0$ (all three terms are used jointly).

Every step computes $\mathcal{L}_{\text{CE}}(x,y)$ on real images. When a matched synthetic $\tilde{x}$ is present, we additionally apply $\mathcal{L}_{\text{PAL}}$, $\mathcal{L}_{\text{InfoNCE}}$, and $\mathcal{L}_{\text{OT}}$ using the same caption $y$. Gradients update only the bridge and decoder; the vision encoder remains frozen. This keeps language supervision tied to real imagery while synthetic data refine the visual representation precisely where the caption attends.

\subsection{Vision Tower: Frozen MaxViT and Why Freezing Helps}
\label{ssec:maxvit-frozen}

The MaxViT backbone (Sec.~\ref{sec:arch}) supplies rich multi-scale features through its interleaved window and grid attention hierarchy. We use only its final, and optionally penultimate, feature maps to extract patch descriptors and do not allow gradients to flow into this tower. Freezing the vision tower is crucial because Bengali captioning data are scarce and often contain synthetic textures; fine-tuning the backbone would overfit to these artifacts and destabilize learning. Keeping MaxViT fixed preserves its strong ImageNet-trained prior while letting the rest of the model adapt to the Bengali language space. It also reduces compute and memory overhead, enabling longer sequences and larger batches during training. This design isolates all alignment pressure onto the projected features consumed by the decoder, preventing the base visual representations from drifting. Only the linear projection layer that maps vision features to the decoder width and the decoder itself are trainable, while cross-attention operates fully in the decoder’s feature space.

\subsection{Language Path: mBART-50 Fine-Tuning for Bengali}
\label{ssec:mbart-training}

The frozen MaxViT outputs are flattened into patch tokens and linearly projected with LayerNorm to the decoder’s hidden width. These projected tokens serve as the key–value memory for an mBART-50 decoder, which autoregressively generates Bengali captions using cross-attention. We enforce the Bengali BOS token (\texttt{bn\_IN}) and keep only the projection and decoder trainable, ensuring the model learns to interpret stable visual features directly in Bengali without altering the vision backbone. Cross-attention maps from the last few decoder layers provide token-level saliency over patches; we reuse them to form weights for the Patch-Alignment Loss, aligning only the regions the decoder attends to. This allows alignment pressure to follow the text’s focus while cross-entropy drives fluent decoding. For stability under mixed-precision training, the bridge layers use the decoder’s dtype. Training minimizes the negative log-likelihood of the Bengali caption under teacher forcing, with PAD tokens masked from the loss:
\begin{equation}
\mathcal{L}_{\text{CE}}(x,y)\;=\;-\sum_{t=1}^{T}\log p_{\theta}\!\big(y_t\,\big|\,y_{<t},\,E(x)\big).
\end{equation}
We gradually unfreeze the decoder layers during training to prevent early overfitting, use gradient checkpointing for memory efficiency, apply AdamW with a One-Cycle learning rate schedule, and clip gradients to stabilize updates. At inference, captions are generated with beam search using a forced Bengali BOS and standard decoding constraints such as length penalty and no-repeat n-grams.

\subsection{Cross-Attention Patch Alignment with Contrastive and Transport Coupling}
\label{ssec:pal}

Decoder cross-attention supplies a token-time saliency over image patches. We average the last \(K\) layers and heads, mask PAD steps, apply a temperature \(\tau\), and optionally keep the top \(\rho\) mass to suppress background, yielding text-conditioned patch weights

\vspace{-0.7cm}
\begin{equation}
\begin{aligned}
w=\mathrm{TopKSoftmax}\!\big(&\tfrac{1}{K}\!\!\sum_{\ell=L-K+1}^{L}\mathrm{mean\_heads}(\mathcal{A}^{(\ell)})\,,\\
& \tau,\rho\big)\in\Delta^{S-1}.
\end{aligned}
\end{equation}

Let \(E(x)=[E_s(x)]_{s=1}^{S}\in\mathbb{R}^{S\times D}\) be Linear+LN projected patch tokens from the frozen encoder. Text-weighted pooling produces caption-relevant descriptors
\begin{equation}
r=\sum_{s}w_s\,E_s(x),\qquad \tilde r=\sum_{s}w_s\,E_s(\tilde x),
\end{equation}
and the patch-alignment loss is
\begin{equation}
\mathcal{L}_{\mathrm{PAL}}(x,\tilde x,y)=1-\cos\!\big(r,\tilde r\big).
\end{equation}

\paragraph{InfoNCE (Global Discrimination).}
Over a minibatch of \(B\) triplets, form the set \(Z=\{r_i,\tilde r_i\}_{i=1}^{B}\) (L2-normalized). Each vector \(z\in Z\) has a single positive \(z^{+}\) (its paired real/synthetic counterpart) and \(2B-2\) negatives. With temperature \(t\),
\begin{equation}
\mathcal{L}_{\mathrm{InfoNCE}}
= -\frac{1}{|Z|}\sum_{z\in Z}
\log
\frac{\exp(\cos(z,z^{+})/t)}{\sum_{z'\in Z\setminus\{z\}}\exp(\cos(z,z')/t)}.
\end{equation}
This sharpens instance-level separation so that pooled descriptors from different captions remain distinct while paired real–synthetic features coincide.

\paragraph{Entropic OT (Fine-Grained Matching).}
Use the same attention weights as masses on patches. Let \(a=\mathrm{top}\text{-}\rho(w)\in\Delta^{S-1}\) and \(b=\mathrm{top}\text{-}\rho(w)\in\Delta^{S-1}\) for \(x\) and \(\tilde x\) (renormalized), and define the cosine cost between patch tokens
\begin{equation}
C_{st}=1-\cos\!\big(E_s(x),E_t(\tilde x)\big).
\end{equation}
The entropic OT objective with regularization \(\varepsilon\) is
\begin{equation}
\mathcal{L}_{\mathrm{OT}}
=\min_{P\in\mathbb{R}^{S\times S}_{\ge 0}}
\;\langle P,C\rangle - \varepsilon\,H(P)
\quad\text{s.t.}\quad
P\mathbf{1}=a,\; P^{\!\top}\mathbf{1}=b,
\end{equation}
solved by Sinkhorn iterations; we use \(\langle P^{\star},C\rangle\) as the loss. OT encourages sparse, location-aware correspondences among caption-attended patches, complementing PAL’s pooled view.

\paragraph{How They Work Together.}
PAL localizes learning to the patches the decoder uses, aligning real and synthetic evidence at the descriptor level. InfoNCE prevents pooled descriptors from drifting toward an easy but uninformative common mode by contrasting across captions, which stabilizes PAL optimization. OT pushes alignment down to patch-to-patch transport under the same text weights, capturing multi-instance or small objects that a single pooled vector may blur. The joint PAL+InfoNCE+OT objective therefore couples caption-time relevance, instance discrimination, and fine-grained spatial matching within one training loop.

%% file: sec/experiments.tex
\section{Experiments \& Results}
\label{sec:experiments}

\subsection{Setup-at-a-Glance: Data, Compute, and Protocols}
\label{ssec:setup}

\noindent\textbf{Datasets.}
We train on 110,000 synthetic images paired with MSCOCO real images and verified Bengali captions. We evaluate on two separate 1k-image sets: MSCOCO-1k for in-domain testing and Flickr30k-1k for out-of-domain testing.

\medskip\noindent\textbf{Compute.}
All experiments use one NVIDIA Tesla P100 GPU with 16 GB memory on Kaggle~\cite{kaggle}. We use mixed precision, gradient accumulation, image size $224^2$, and seed $42$.

\medskip\noindent\textbf{Model.}
The model uses a frozen MaxViT encoder and an mBART-50 decoder. A linear+LayerNorm bridge connects the two modules. Cross-attention from the last two decoder layers is used for PAL, InfoNCE, and OT.

\medskip\noindent\textbf{Training.}
We train for 7 epochs with AdamW~\cite{loshchilov2019decoupled}, One-Cycle LR, mixed precision, and gradient clipping. The encoder stays frozen, while the decoder is progressively unfrozen.

\medskip\noindent\textbf{Loss settings.}
We use $\lambda_{\text{PAL}}{=}0.5$, $\tau{=}1.0$, $\rho{=}0.5$, InfoNCE temperature $t{=}0.07$, and OT regularization $\varepsilon{=}0.05$ with 30 Sinkhorn iterations. The loss weights are $\alpha{=}0.3$ and $\beta{=}0.5$.

\medskip\noindent\textbf{Metrics.}
We report BLEU-1/2/3/4~\cite{papineni2002bleu}, METEOR~\cite{lavie2007meteor}, and BERTScore-F1~\cite{zhang2020bertscore}.

\subsection{Positioning Against Representative Captioning Baselines}
\label{ssec:baselines}

To contextualize our approach, we re-implemented and trained two representative captioning baselines from prior work under our setup and dataset: a ViT encoder with a Bangla GPT-2 decoder \cite{ved}, and a CNN+LSTM captioner \cite{bornon2021bengali}. Their performance on Flickr30k-1k is reported in \Cref{tab:baseline-comparison}, alongside our model that combines PAL, InfoNCE, and OT.

\begin{table}[t]
\centering
\setlength{\tabcolsep}{4pt}
\renewcommand{\arraystretch}{1.2}
\caption{Flickr30k-1k (out-of-domain): Comparison of our model (PAL+InfoNCE+OT) with established captioners. B1–B4 are BLEU-1 to BLEU-4. Higher is better.}
\label{tab:baseline-comparison}
\scriptsize
\begin{tabular}{lcccccc}
\toprule
Method & B1 & B2 & B3 & B4 & METEOR & BERT-F1 \\
\midrule
Our model  & \textbf{49.14} & \textbf{29.64} & \textbf{18.11} & \textbf{12.29} & 27.98 & 71.20 \\
ViT + Bangla GPT-2 \cite{ved} & 26.84 & 12.32 & 4.68 & 3.21 & 26.62 & 57.34 \\
CNN + LSTM \cite{bornon2021bengali} & 28.03 & 15.45 & 6.51 & 3.98 & 22.57 & 70.98 \\
Google PaliGemma (3B) \cite{beyer2024paligemma} & 35.40 & 22.19 & 15.81 & 11.28 & \textbf{32.01} & \textbf{77.96} \\
mBLIP mT0-XL \cite{geigle2023mblip} & 36.73 & 24.17 & 14.45 & 9.53 & 28.92 & 76.92 \\
InternVL 2.5 (4B) \cite{chen2024expanding,chen2024far,chen2024internvl,gao2024mini} & 18.71 & 5.64 & 3.14 & 2.13 & 23.35 & 69.98 \\ 
\bottomrule
\end{tabular}
\end{table}

As shown in \Cref{tab:baseline-comparison}, both representative captioning baselines underperform our model despite using the same data and compute. ViT+GPT-2 achieves competitive METEOR but much lower BERTScore and BLEU scores, indicating weaker lexical and semantic agreement with the references. CNN+LSTM reaches similar BERTScore but trails on BLEU-4 and METEOR, suggesting limited lexical diversity. Our model surpasses both across BLEU-1–4 and METEOR while maintaining strong BERTScore, demonstrating more accurate and fluent Bengali captions.

\subsection{Do We Caption Better? (Main Results)}
\label{ssec:main-results}

We evaluate one model on disjoint MSCOCO-1k (in-domain) and Flickr30k-1k (out-of-domain). \Cref{tab:main-ours-compact} reports corpus-level scores.

\begin{table}[t]
\centering
\scriptsize
\setlength{\tabcolsep}{5pt}
\renewcommand{\arraystretch}{1.2}
\caption{Single-model performance (PAL + InfoNCE + OT) on MSCOCO-1k (in-domain) and Flickr30k-1k (out-of-domain). B1–B4 are BLEU-1 to BLEU-4.}
\label{tab:main-ours-compact}
\begin{tabular}{lcccccc}
\toprule
Dataset & B1 & B2 & B3 & B4 & METEOR & BERT-F1 \\
\midrule
MSCOCO-1k    & 38.61 & 25.11 & 17.43 & 12.00 & 28.14 & 75.40 \\
Flickr30k-1k & 49.14 & 29.64 & 18.11 & 12.29 & 27.98 & 71.20 \\
\bottomrule
\end{tabular}
\end{table}

On MSCOCO-1k the model reaches BLEU-4 of 12.00 with strong semantic similarity (BERTScore-F1 75.40, METEOR 28.14), indicating fluent Bengali captions that closely track the reference descriptions. Crucially, on the shifted Flickr30k-1k domain BLEU-4 remains comparable at 12.29 and METEOR is stable, while BERTScore-F1 drops moderately to 71.20. The increase in BLEU-1 (38.61 to 49.14) is consistent with longer references and higher unigram overlap in Flickr30k. Together, these trends show the model transfers beyond its training domain: PAL ties the alignment term to the patches the decoder attends to, and the complementary InfoNCE and OT terms preserve instance discrimination and fine-grained correspondences.

\subsection{What Makes It Tick? Compact Ablations with Insights}
\label{ssec:ablations}

We compare different training variants on Flickr30k-1k. Table~\ref{tab:ablations-flickr} shows the result.

\begin{table}[t]
\centering
\caption{Ablation results on Flickr30k-1k. Higher is better.}
\label{tab:ablations-flickr}
\scriptsize
\begin{tabular}{lcccccc}
\hline
Method & B1 & B2 & B3 & B4 & METEOR & BERT-F1 \\
\hline
CE (Real-only) & 36.57 & 23.81 & 12.37 & 5.80 & 24.96 & 68.38 \\
CE (Real+Syn) & 40.34 & 24.46 & 12.93 & 5.91 & 24.51 & 68.69 \\
CE + InfoNCE & 43.60 & 25.67 & 11.98 & 7.50 & 24.99 & 69.70 \\
CE + InfoNCE + OT & 42.56 & 24.00 & 11.85 & 7.52 & 26.22 & 70.21 \\
PAL & 46.53 & 27.80 & 16.30 & 10.19 & 27.29 & 70.97 \\
PAL + InfoNCE & 46.86 & 27.96 & 16.57 & 10.34 & 27.49 & 70.69 \\
\textbf{PAL + InfoNCE + OT} & \textbf{49.14} & \textbf{29.64} & \textbf{18.11} & \textbf{12.29} & \textbf{27.98} & \textbf{71.20} \\
\hline
\end{tabular}
\end{table}

Adding synthetic images to CE gives only small gains. BLEU-4 changes from 5.80 to 5.91, so synthetic data alone is not enough. Adding InfoNCE and OT improves BLEU-4 to about 7.5, but the scores are still below the PAL-based models.

PAL gives the main improvement. It raises BLEU-4 from 7.52 to 10.19 and improves METEOR and BERTScore. The full model performs best on all metrics, reaching 12.29 BLEU-4, 27.98 METEOR, and 71.20 BERTScore-F1.

These results show that PAL is the key part. InfoNCE improves instance separation, and OT adds patch-level matching. Together, they make synthetic images more useful for Bengali captioning.

\subsection{Sensitivity to PAL Hyperparameters}
\label{sec:pal-hparam}

We analyze how PAL behaves under different pooling hyperparameters: the loss weight $\lambda_{\text{PAL}}$, attention temperature $\tau$ (which sharpens or smooths the last-$K$ decoder cross-attention maps), and the top-$k$ ratio $\rho$ (fraction of attended patch mass retained; see Eq.~(8)–(9) in §4.6).  
This experiment was run on a smaller subset of the data and with fewer epochs (6), so the absolute scores are lower than our main results (§5.3), but trends are consistent.

\begin{table}[h]
\centering
\caption{PAL hyperparameter sensitivity on a reduced-data 6-epoch run. Relative trends guide our chosen configuration.}
\small
\setlength{\tabcolsep}{6pt}
\begin{tabular}{ccccccc}
\toprule
$\lambda$ & $\tau$ & $\rho$ & BLEU-4 & METEOR & BERT-F1 \\
\midrule
0.3 & 0.7 & 0.5 & 2.77 & 12.8 & 62.2 \\
0.5 & 0.7 & 0.1 & 2.50 & 17.5 & 63.7 \\
0.5 & 1.0 & 0.5 & 2.37 & 10.9 & 61.3 \\
0.5 & 1.3 & 0.5 & 2.42 & 15.8 & 61.9 \\
\bottomrule
\end{tabular}
\label{tab:pal-hparam}
\end{table}

Lower $\tau$ (sharper attention) boosts lexical precision (BLEU-4, METEOR), while higher $\tau$ spreads weights and supports broader coverage. Small $\rho$ focuses on top evidence patches (higher METEOR/BERT-F1), whereas larger $\rho$ includes more context and stabilizes BLEU-4. Moderate $\lambda{=}0.5$ balances these effects.  
We adopt $\mathbf{\lambda{=}0.5,\ \tau{=}1.0,\ \rho{=}0.5}$ for our main experiments, as it offers stable alignment behavior while maintaining strong precision and semantics.

\subsection{Learning Where the Words Look: Convergence and Alignment}
\label{ssec:dynamics}

\begin{figure}[t]
\centering
\includegraphics[width=0.48\linewidth]{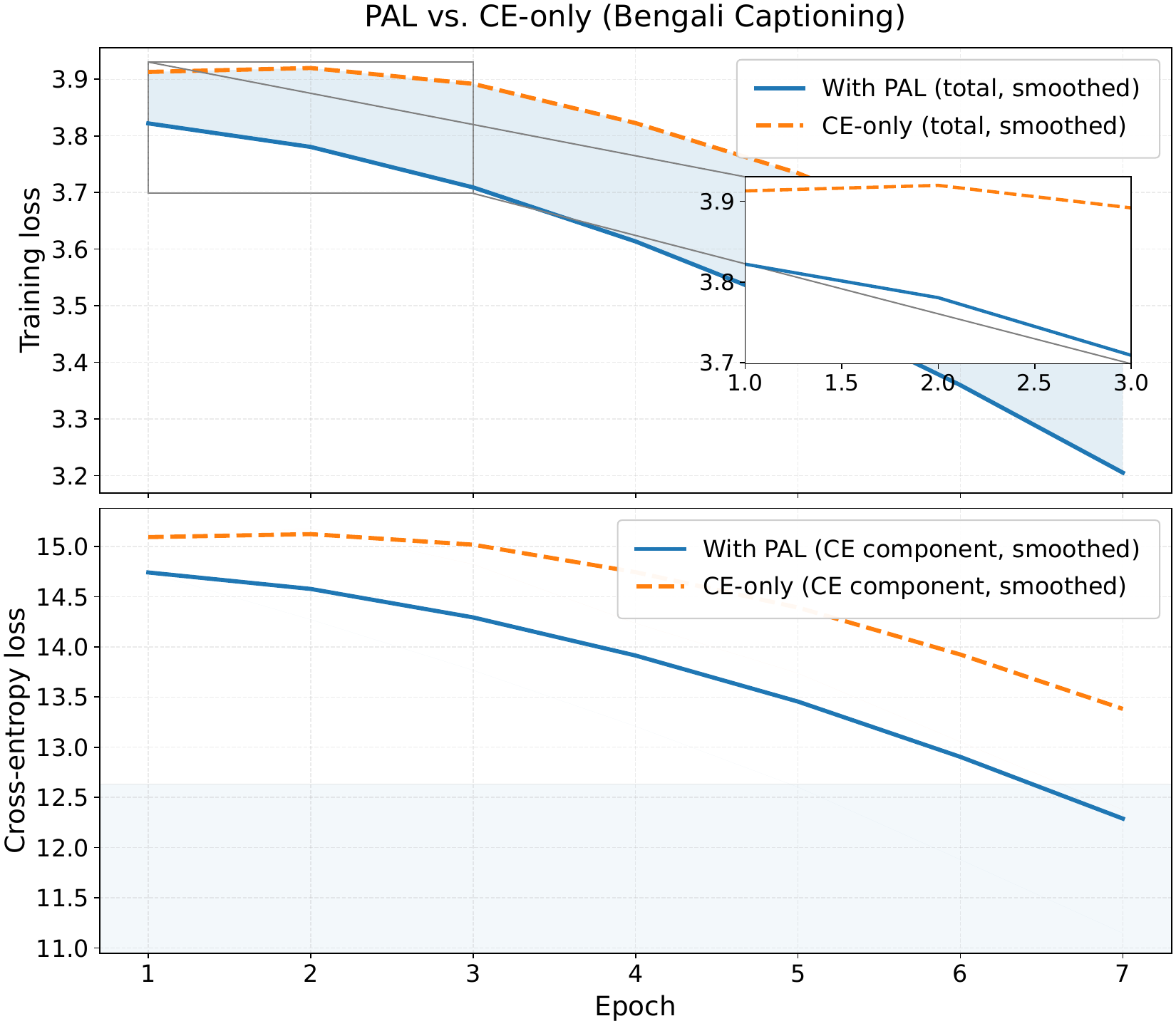}
\hfill
\includegraphics[width=0.48\linewidth]{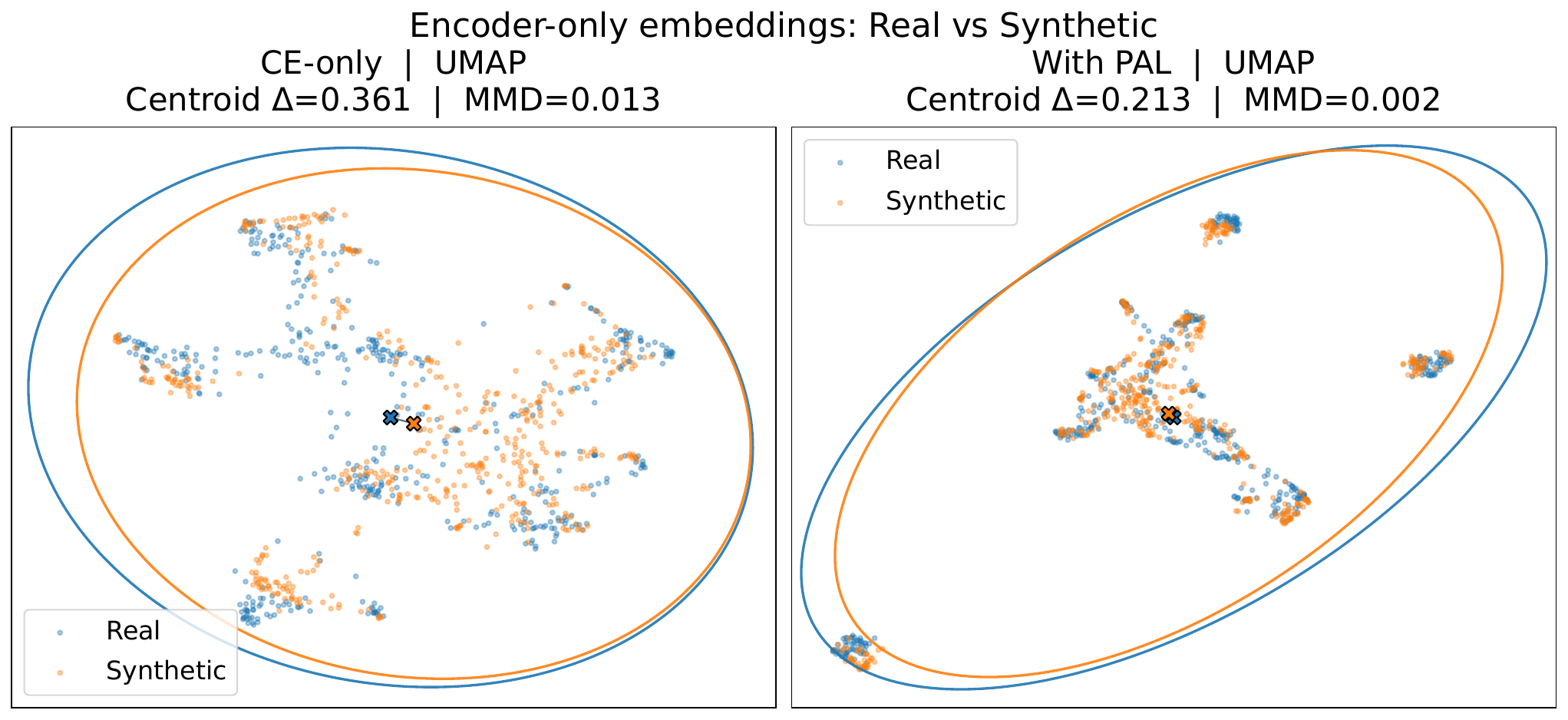}
\caption{PAL improves optimization and alignment. Left: PAL reduces total and CE loss. Right: PAL reduces the real--synthetic centroid gap from 0.361 to 0.213 and MMD-RBF from 0.013 to 0.002.}
\label{fig:loss-and-embed}
\end{figure}

We compare PAL against a captioning-only objective and observe faster convergence and lower final losses for PAL in \Cref{fig:loss-and-embed} (left). The cross-attention-guided pooling directs gradients onto tokens linked to caption-relevant patches, which reduces the caption loss itself rather than acting as a mere regularizer.

The embedding view in \Cref{fig:loss-and-embed} (right) further shows how PAL narrows the domain gap where it matters. Real and synthetic patch descriptors move closer in the regions the decoder actually attends to, with centroid distance shrinking from 0.361 to 0.213 and Maximum Mean Discrepancy with a Radial Basis Function kernel decreasing from 0.013 to 0.002. The distribution does not collapse; overlap increases around text-relevant evidence while background variation remains.

These dynamics indicate that PAL teaches the model where to align. By reweighting patches with decoder cross-attention and aligning pooled descriptors of real and synthetic pairs, it improves real-synthetic agreement in attended regions and stabilizes learning under limited Bengali supervision.

%% file: sec/llms.tex
\section{Use of Large Language Models (LLMs)}
\label{sec:llms}
While writing the paper, we used AI assistance for polishing a few sentences and for some minor debugging of the code. The authors remain fully
responsible for both the manuscript and the code.

%% file: sec/conclusion.tex
\section{Conclusion}
\label{sec:conclusion}

We presented a compute-aware Bengali image captioning framework that combines Patch-Alignment Loss (PAL), InfoNCE, and Sinkhorn-based OT to improve real-synthetic patch alignment under scarce paired supervision. PAL aligns decoder-attended real and synthetic patches, while InfoNCE and OT strengthen global discrimination and fine-grained correspondence. Experiments on MSCOCO-1k and Flickr30k-1k show consistent gains in BLEU-$n$, METEOR, and BERTScore, demonstrating that targeted real--synthetic alignment can produce more accurate and generalizable Bengali captions. Future work may extend this framework with region-level cues, Bengali CLIP-style priors, or compact adapters for stronger low-resource vision--language modeling.